\documentclass{article}
\usepackage{ijcai21}

\usepackage{times}

\usepackage{soul}
\usepackage{url}
\usepackage[hidelinks]{hyperref}
\usepackage[utf8]{inputenc}
\usepackage[small]{caption}
\usepackage{graphicx}
\usepackage{amsmath}
\usepackage{booktabs}
\usepackage{float}
\usepackage{enumerate}
\usepackage{algorithm}
\usepackage{algorithmic} 
\usepackage{multirow}
\usepackage{multicol}
\usepackage{arydshln}
\usepackage{subfigure}
\usepackage{graphicx}
\title{GFL: A Decentralized Federated Learning Framework Based On Blockchain}

\author{

Yifan Hu$^1$\and
Yuhang Zhou$^2$\and
Jun Xiao$^{3}$\and
Chao Wu$^4$\footnote{Contact Author}\\
\affiliations
$^1$Polytechnic Institute, Zhejiang University\\
$^2$School of Software Technology, Zhejiang University\\
$^3$College of Computer Science and Technology, Zhejiang University\\
$^4$School of Public Affairs, Zhejiang University\\
\emails
\{yifan\_hu, yuhang\_zhou, chao.wu\}@zju.edu.cn,
 junx@cs.zju.edu.cn
}

\begin{document}

\maketitle

\begin{abstract}
 Federated learning(FL) is a rapidly growing field and many centralized and decentralized FL frameworks have been proposed. However, it is of great challenge for current FL frameworks to improve communication performance and maintain the security and robustness under malicious node attacks. In this paper, we propose Galaxy Federated Learning Framework(GFL), a decentralized FL framework based on blockchain. GFL introduces the consistent hashing algorithm to improve communication performance and proposes a novel ring decentralized FL algorithm(RDFL) to improve decentralized FL performance and bandwidth utilization. In addition, GFL introduces InterPlanetary File System(IPFS) and blockchain to further improve communication efficiency and FL security. Our experiments show that GFL improves communication performance and decentralized FL performance under the data poisoning of malicious nodes and non-independent and identically distributed(Non-IID) datasets.
\end{abstract}

\section{Introduction}

In the past few years, many federated learning(FL) frameworks have been proposed. Conventional FL framework \cite{ryffel2018generic,caldas2018leaf} uses a highly centralized architecture where a centralized node collects gradients or model parameters from data nodes to update the global model. In a real FL scenario, the centralized node of the conventional FL framework suffers from communication pressure and communication bandwidth bottleneck \cite{tang2020communication,philippenko2020artemis}. Moreover, communication pressure brings challenge to the stability of the centralized node.

To avoid the problems caused by the centralized FL framework, the decentralized FL framework \cite{he2019central,li2020blockchain,hu2019decentralized} has been proposed. The decentralized FL framework removes the centralized node and synchronizes FL updates among the data nodes, then performs aggregation. 


Compare with the centralized FL framework, the decentralized FL framework has made some progress in breaking communication bottlenecks and improving FL stability. However, It still faces challenges in communication pressure, decentralized FL performance and security \cite{zhao2019mobile,li2020blockchain}. The additional communication caused by the decentralized FL framework to synchronize FL updates causes more communication pressure. In addition, due to the data poisoning from malicious nodes \cite{li2020blockchain} in the data nodes and the problem of not independent and identically distributed(Non-IID) \cite{zhao2018federated}, the aggregation algorithm used in the existing decentralized FL framework cannot achieve competitive performance. Moreover, the updating process of existing decentralized FL framework is suffering security risk of tampering and the source of FL updates cannot be traced due to the lack of consensus mechanisms.

To tackle aforementioned problems, a blockchain-based decentralized FL framework Galaxy Federated Learning Framework(GFL) is proposed in this paper. Consistent hashing algorithm \cite{lamping2014fast} is employed in GFL to construct a ring topology of data nodes, which aims to reduce the communication pressure and improve topology stability. Besides, an innovative ring decentralized FL algorithm(RDFL) is designed in the proposed GFL. Ring-allreduce\footnote{https://andrew.gibiansky.com/blog/machine-learning/baidu-allreduce/} algorithm and knowledge distillation \cite{hinton2015distilling} are introduced in RDFL to benefit the bandwidth utilization and decentralized FL performance. Additionally, InterPlanetary File System(IPFS) \cite{benet2014ipfs} and blockchain have been introduced to further improve the communication efficiency and security of decentralized FL.

In summary, the main contributions of GFL are as follows:
\begin{itemize}

\item[1)] A new data node topology mechanism for decentralized FL has been designed in this work where consistent hashing algorithm is employed. The proposed mechanism is able to significantly reduce communication pressure and improve topology stability. To the best of our knowledge, this is the first attempt to introduce the consistent hashing algorithm for data node topology design in decentralized FL framework.

\item[2)] A novel ring decentralized federated learning(RDFL) algorithm designed to improve the bandwidth utilization and decentralized FL performance in training.

\item[3)] To improve the communication performance and security of the decentralized FL framework, we introduce IPFS technology to reduce system communication pressure and blockchain to improve the FL security .

\end{itemize}

\section{Related Work}

Federated learning(FL) was first proposed by Google \cite{mcmahan2017communication} and was widespread by Google's blog post\footnote{https://ai.googleblog.com/2017/04/federated-learning-collaborative.html}. Google \cite{mcmahan2017communication} proposes a FL process that collects locally calculated gradients and aggregates them at the central node. To help build FL task, some centralized FL frameworks have been proposed. Representatives of these frameworks are FATE\footnote{https://fate.fedai.org/}, TensorFlow-Federated(TFF)\footnote{https://www.tensorflow.org/federated}, PaddleFL\footnote{https://github.com/PaddlePaddle/PaddleFL}, LEAF \cite{caldas2018leaf} and PySyft \cite{ryffel2018generic}. However, these centralized FL frameworks still have the problem of communication bottleneck and stability.

To avoid the problems caused by centralized FL framework, the research on the decentralized FL framework has attracted much attention. Hu \cite{hu2019decentralized} proposed a decentralized FL algorithm based on Gossip algorithm and model segmentation, Roy \cite{roy2019braintorrent} proposed a peer-to-peer decentralized FL algorithm, Lalitha \cite{lalitha2018fully} explored fully decentralized FL algorithm and Li \cite{li2020blockchain} proposed a blockchain-based decentralized FL framework. However, the decentralized FL framework still faces challenges in the decentralized FL performance and security. 

To tackle the communication problem of decentralized FL framework, current research focuses on researching novel communication compression or model compression techniques to reduce the communication pressure. For example, Hu \cite{hu2019decentralized} utilizes the gossip algorithm to improve bandwidth utilization and model segmentation to reduce communication pressure. Amiri \cite{amiri2020federated} and Konečný \cite{konevcny2016federated} propose model quantification methods to reduce communication pressure. Tang \cite{tang2018communication} and Koloskova \cite{koloskova2019decentralized} introduce communication compression methods to reduce communication pressure. 

To improve the FL performance on Non-IID dataset, existing research mainly focuses on centralized FL. Sharing datasets \cite{zhao2018federated} and knowledge distillation \cite{jeong2018communication,itahara2020distillation} are all novel algorithms. In decentralized FL scenario, to improve the decentralized FL performance on Non-IID dataset and data poisoning of malicious data nodes, Li \cite{li2020blockchain} proposed a decentralized FL framework, but did not achieve better performance than \textit{Federated Averaging}(FedAvg) \cite{mcmahan2017communication} on the situation that without data poisoning.

To further protect the data privacy of FL, existing research focuses on new defense methods, such as differential privacy \cite{geyer2017differentially} and multi-party secure computing \cite{melis2019exploiting}. For decentralized FL framework, there is also research on applying blockchain technology to decentralized FL to improve the security \cite{lu2019blockchain,kim2019blockchained,li2020blockchain}.






\section{The Proposed Framework}
In this section, we firstly describe how the designed topology mechanism in GFL utilizing consistent hashing algorithm to build a ring decentralized FL topology for data nodes. Then we describe RDFL algorithm in GFL. Finally, we describe how GFL utilizes IPFS and blockchain to reduce communication pressure and improve FL security.

\subsection{Ring Decentralized FL Topology}

Consider a group of $n$ data nodes among which there are $m$ trusted data nodes and $n-m$ untrusted data nodes. These $n$ data nodes are represented by the symbol $\mathcal{D}=\{DP_1, DP_2, DP_3,...,DP_n\}$. 

\begin{figure}[h]
    \centering
     \includegraphics[width=0.4\textwidth]{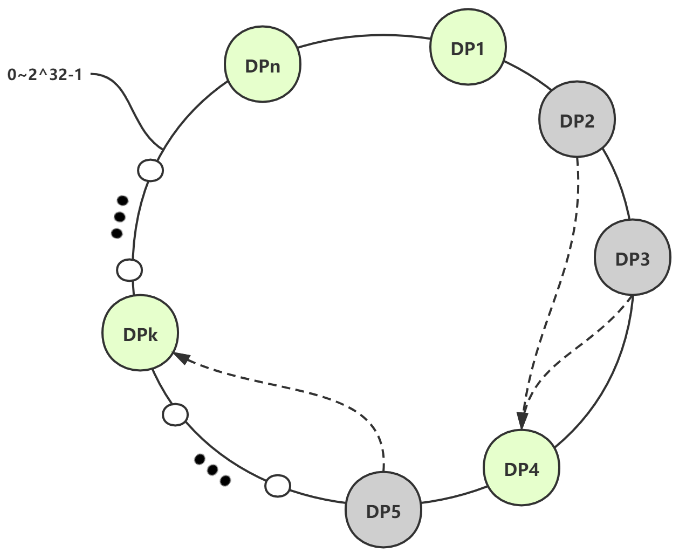}
    \caption{Ring Topology Without Virtual Nodes}
    \label{fig:ring_no_virtual_nodes}
\end{figure}

GFL utilizes consistent hashing algorithm to construct a ring topology of $n$ data nodes. The consistent hash value $H_{k}=Hash(DP_{k}^{ip})\subseteq[0,2^{32}-1]$, $DP_{k}^{ip}$ represents the ip of $DP_k, k\subseteq[1,n]$. Data nodes are distributed on the ring with a value range $[0,2^{32}-1]$ according to the consistent hash value. The untrusted data nodes will send local models to the nearest trusted data node found on the ring topology in a clockwise direction. Figure \ref{fig:ring_no_virtual_nodes} shows the ring topology constructed by the consistent hashing algorithm. The green data nodes represent trusted data nodes and the gray data nodes represent untrusted data nodes. According to the clockwise principle, untrusted data nodes $DP_2$ and $DP_3$ send models to trusted data provider $DP_4$. Untrusted data node $DP_5$ sends models to the nearest trusted data node $DP_k$.

\begin{figure}[h]
    \centering
      \includegraphics[width=0.4\textwidth]{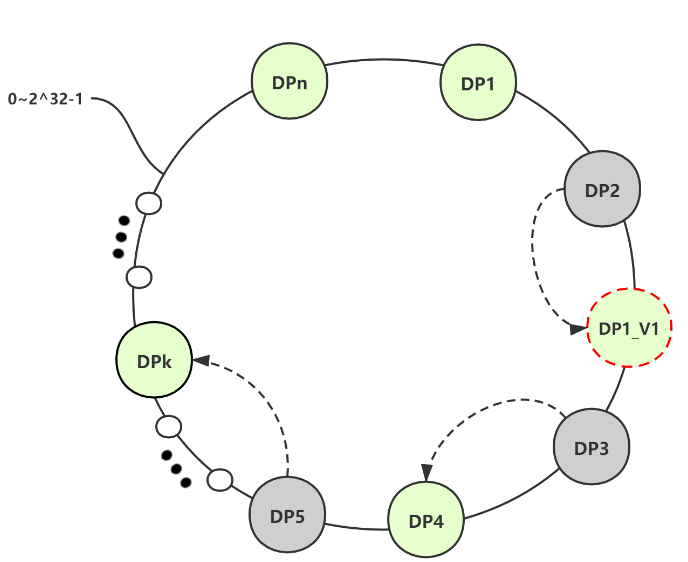}
    \caption{Ring Topology With Virtual Nodes}
    \label{fig:ring_with_virtual_nodes}
\end{figure}

With the help of consistent hashing algorithm, different untrusted data nodes send models to different trusted nodes, which reduces the communication pressure of trusted node effectively. To make the distribution of trusted nodes on the ring more uniform, GFL introduces virtual nodes of trusted nodes in the ring topology to further reduce communication pressure. Figure \ref{fig:ring_with_virtual_nodes} shows a ring topology with virtual nodes. The green nodes with red dashed lines represent virtual nodes. $DP_{1}^{v1}$ is the virtual node of $DP_1$. If the untrusted data nodes find that the closest trusted node is a virtual node in a clockwise direction, they will directly send the models to the trusted node corresponding to the virtual node.


\subsection{RDFL Algorithm}
\begin{figure}[h]
    \centering
    \includegraphics[width=0.5\textwidth]{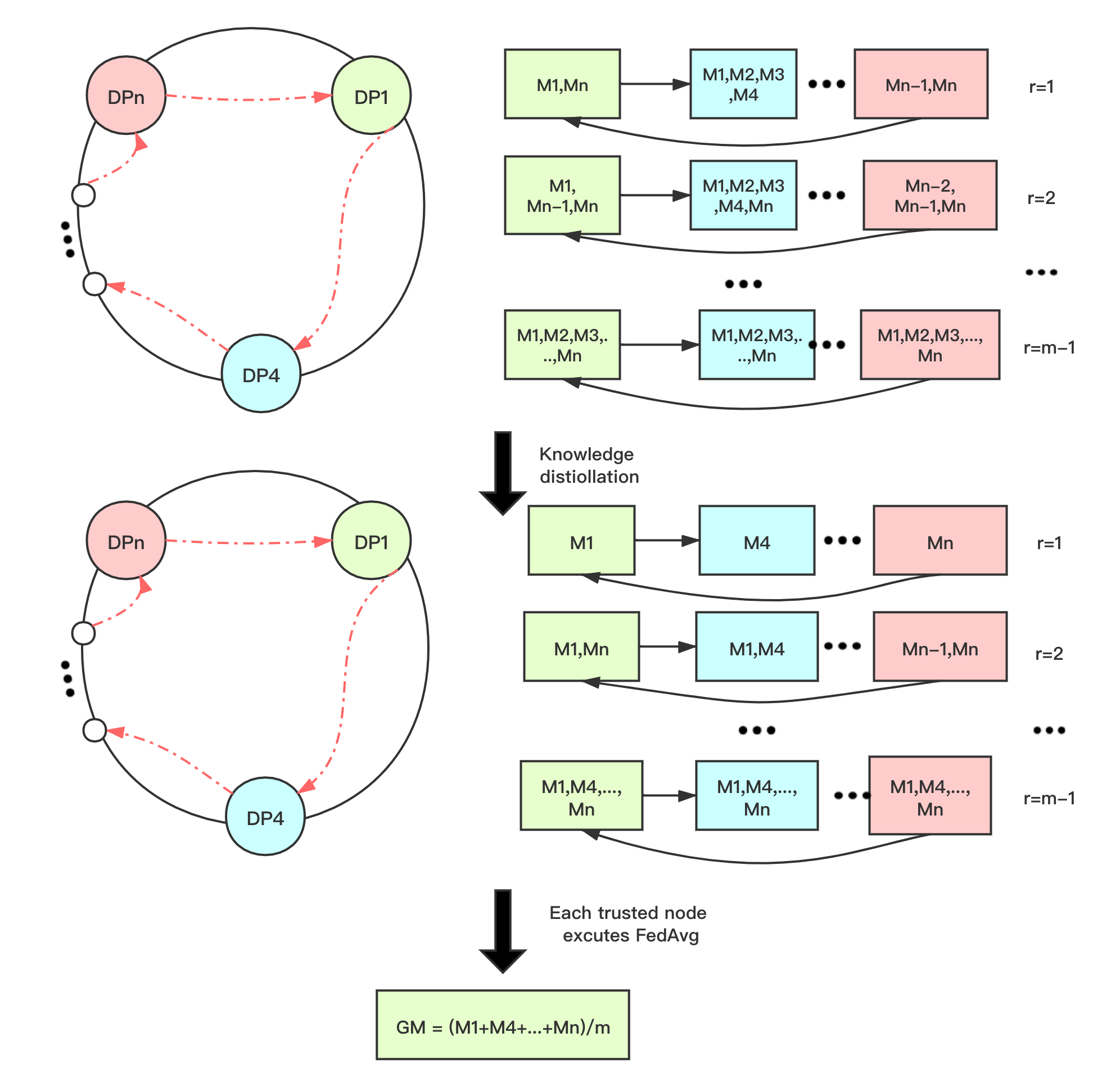}
    \caption{Ring Decentralized Federated Learning}
    \label{fig:rdfl}
\end{figure}

Based on the ring decentralized topology constructed by the consistent hash algorithm, GFL starts to execute the RDFL algorithm. The trusted node follows the RDFL algorithm process as illustrated in Figure \ref{fig:rdfl}. $M_{k}$ represents the model of data node $DP_{k}$. $r$ represents the number of rounds to execute models synchronization and $m$ represents the number of trusted nodes. At each iteration, 
\textbf{(1)}each data node loads the global model for local training and synchronizes its models in a clockwise direction, \textbf{(2)}utilizes knowledge distillation to converge the dark knowledge of other models owned by the trusted node to the local model and synchronize local model among trusted nodes, \textbf{(3)}each trusted node executes \textit{Federated Averaging}(FedAvg) \cite{mcmahan2017communication} to generate a new global model and starts next iteration. The RDFL algorithm training process is described in Algorithm \ref{alg:algorithm}. Next, we describe each step of the RDFL algorithm in detail.

\textbf{(1) Train and Synchronize Model} Each data node $DP_{k}$ firstly loads the global model $GM$ generated in the previous iteration for local training and obtains a new local model $M_{k}$. Then, the untrusted node sends the local model to the nearest trusted node according to the clockwise principle. Further, the trusted node synchronizes models so that all trusted nodes obtain the model of the remaining trusted nodes. To improve the bandwidth utilization between trusted nodes, RDFL algorithm introduces the Ring-allreduce algorithm and the clockwise principle in the consistent hash algorithm to perform model synchronization.


It can be seen from Figure \ref{fig:ring_no_virtual_nodes} that the trusted node $DP_{4}$ obtains $M_2$ and $M_3$ from the untrusted nodes $DP_2$ and $DP_3$ before the synchronization starts. After the first round of synchronization($r=1$), $DP_{1}$ has $M_1$ and $M_{n}$, $DP_{4}$ has $M_1$, $M_2$, $M_3$ and$M_4$, $DP_{n}$ has $M_n$ and $M_{n-1}$. After $m-1$ rounds of synchronization($r=m-1$), all trusted nodes have the models of other trusted nodes.


\textbf{(2) Knowledge Distillation} Due to the synchronization, each trusted node obtains the models from the remaining trusted data nodes. Then, each trusted node conducts knowledge distillation \cite{hinton2015distilling} locally. Through knowledge distillation, the dark knowledge in a powerful teacher model can be transferred to the student model with little knowledge. The loss function $L_\text{student}$ of the student model is shown in formula (1).

\begin{align}
& L_{\text {student}}=L_{C E}+D_{K L}\left(p_{\text {teacher}} \| p_{\text {student}}\right)\label{XX}\\
& p_{\text {teacher}}=\frac{\exp (z / T)}{\sum_{i} \exp \left(z_{i} / T\right)}\label{yy}
\end{align}  

$L_{CE}$ and $D_{KL}$ respectively represent cross entropy and Kullback Leibler(KL) divergence. $p$ represents the output of the model after the softmax activation function. $z$ represents the logits output by the model and $T$ is a hyper-parameter representing the temperature of distillation. Knowledge distillation minimizes $L_{student}$ to allow the student model to learn dark knowledge. RDFL utilizes KL divergence to measure the difference in distribution between models. $L_{student}$ of the student model in RDFL is shown in formula (3). $z_{student}$ and $z_{teacher}$ represent the logits of the model.

\begin{align}
& L_{\text {student}}=L_{CE}+D_{KL}\left(z_{\text {teacher}} \| z_{\text {student}}\right)\label{zz}
\end{align}  

In knowledge distillation step of the RDFL algorithm, each trusted node $DP_{k}$ uses the local model $M_k$ as a student and the remaining models owned by $DP_{k}$ will be used as teachers to distill the dark knowledge to $M_k$. To avoid the data poisoning of malicious nodes in untrusted nodes, RDFL only allows the top 30\% models with the smallest KL divergence from $M_k$ as teacher models. With the FL round increases, the number of teacher models will be dynamically increased to improve generalization, but not more than 50\%. Each trusted node only keeps the local model and discards remaining models after knowledge distillation. In Figure \ref{fig:rdfl}, the trusted node $DP_{1}$ retains the local model $M_1$ after distillation, $DP_{4}$ retains $M_4$ and $DP_{n}$ retains $M_n$. Then, the trusted nodes utilize the Ring-allreduce algorithm and the clockwise rule to synchronize the local models of the trusted nodes. After synchronization, all trusted nodes have local models of other trusted nodes.

\textbf{(3) Federated Averaging}
In this step, each trusted node $DP_{k}$ executes FedAvg to obtain a new global model. Then, each trusted node sends the new global model counterclockwise to the untrusted nodes based on the ring topology and starts next iteration.

\begin{algorithm}[tb]
\caption{RDFL algorithm}
\label{alg:algorithm}
\textbf{Input}: The initial global model $GM_{0}$\\
\textbf{Output}: The New global model \\
\textbf{Procedure}: Data Node Executes
\begin{algorithmic}[1] 
\FOR{each FL round $t=1, 2, 3, . . . , T$}
\STATE Each data node $Training(GM_{t-1})$;
\FOR{each untrusted data nodes $DP_{h},h\subseteq\mathcal{D}$ in parallel}
\STATE /*$M_{h}^{t}$ is the local model of  $DP_{h}$ in round $t$*/;
\STATE Send $M_{h}^{t}$ to the nearest trusted node found on the ring topology in a clockwise direction;
\ENDFOR
\STATE  Each trusted node synchronizes its models;
\FOR{each trusted node $DP_{k},k\subseteq\mathcal{D}$ in parallel}
    \STATE /* $\mathcal{M}_{k}^{t}$ is a collection of models owned by $DP_{k}$ in round $t$ */
    \STATE $M_{k}^{t} \leftarrow Distillation(\mathcal{M}_{k}^{t})$ ;
\ENDFOR
\STATE Each trusted node synchronizes its local model;
\STATE /* $m$ is the number of trusted nodes */
\STATE Each trusted node executes FedAvg to obtain new global model $GM_{t}$: $GM_{t} \leftarrow \frac{1}{m} \sum_{i=1}^m M_{i}^{t} $;
\ENDFOR
\end{algorithmic}
\end{algorithm}

\subsection{IPFS and Blockchain}
In the conventional decentralized FL algorithm \cite{he2019central}, the transfer of the models among data nodes occupies a lot of communication overhead and causes serious communication pressure. Moreover, these communication is suffering security risk of tampering and the source of models cannot be traced. 

\textbf{Communication Pressure} To reduce the communication pressure, GFL introduces IPFS as a model storage system. IPFS is composed of trusted nodes in GFL. Files in IPFS will be divided into multiple pieces stored on different nodes and IPFS will generate the IPFS hash corresponding to the file. The IPFS hash is a 46 bytes string and the corresponding file can be obtained from IPFS through the IPFS hash. In GFL, since the model files are stored in pieces in trusted nodes, the risk of data privacy leakage is further reduced.


\textbf{Communication Security} To improve the security of decentralized FL framework, GFL introduces blockchain as a communication system.

In GFL, IPFS hashes sent by data nodes are transmitted in the form of blockchain transactions. Every transaction is verified by the consensus algorithm to reduce the risk of tampering. In addition, since the blockchain records every transaction information, GFL can trace back to malicious nodes based on the transaction information.

It is worth noting that due to the openness of the blockchain, any node on the blockchain can obtain transaction information in the blockchain. To avoid the leakage of the IPFS hash, GFL encrypts the IPFS hash to protect the privacy of data nodes.


\section{Framework Design}
This section describes the architecture and the encryption mechanism of GFL in detail.

\begin{figure}[htb]
    \centering
    \includegraphics[width=0.5\textwidth]{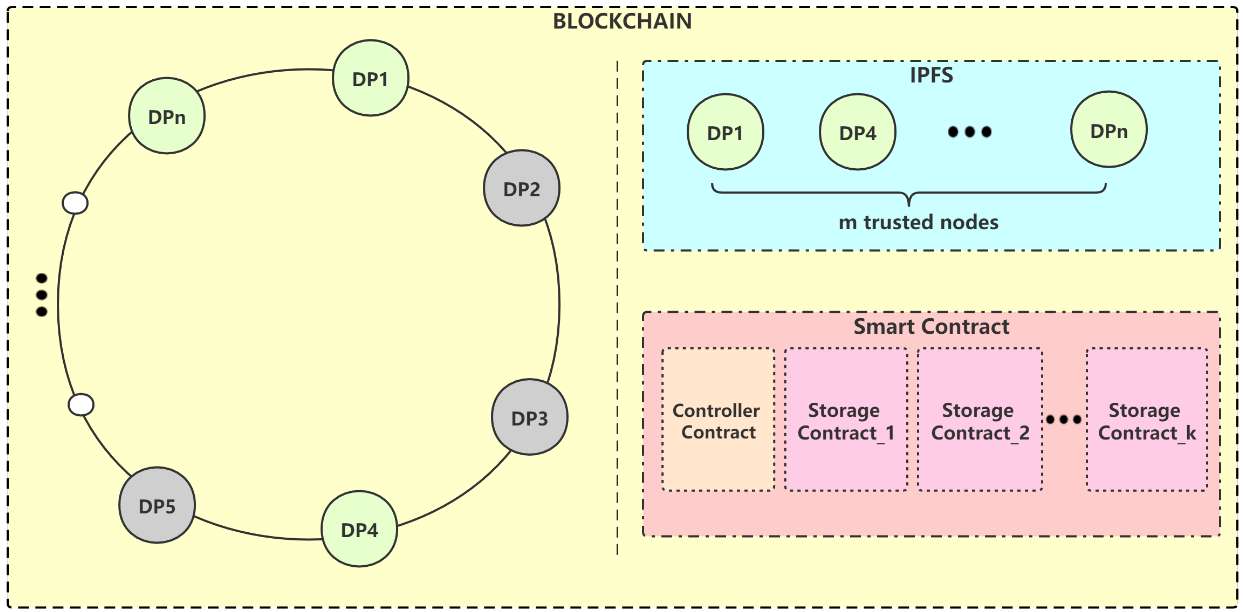}
    \caption{Architecture of GFL}
    \label{fig:gfl_architecture}
\end{figure}

\subsection{Architecture}

\begin{figure*}[htb]
    \centering
    \includegraphics[width=1\textwidth]{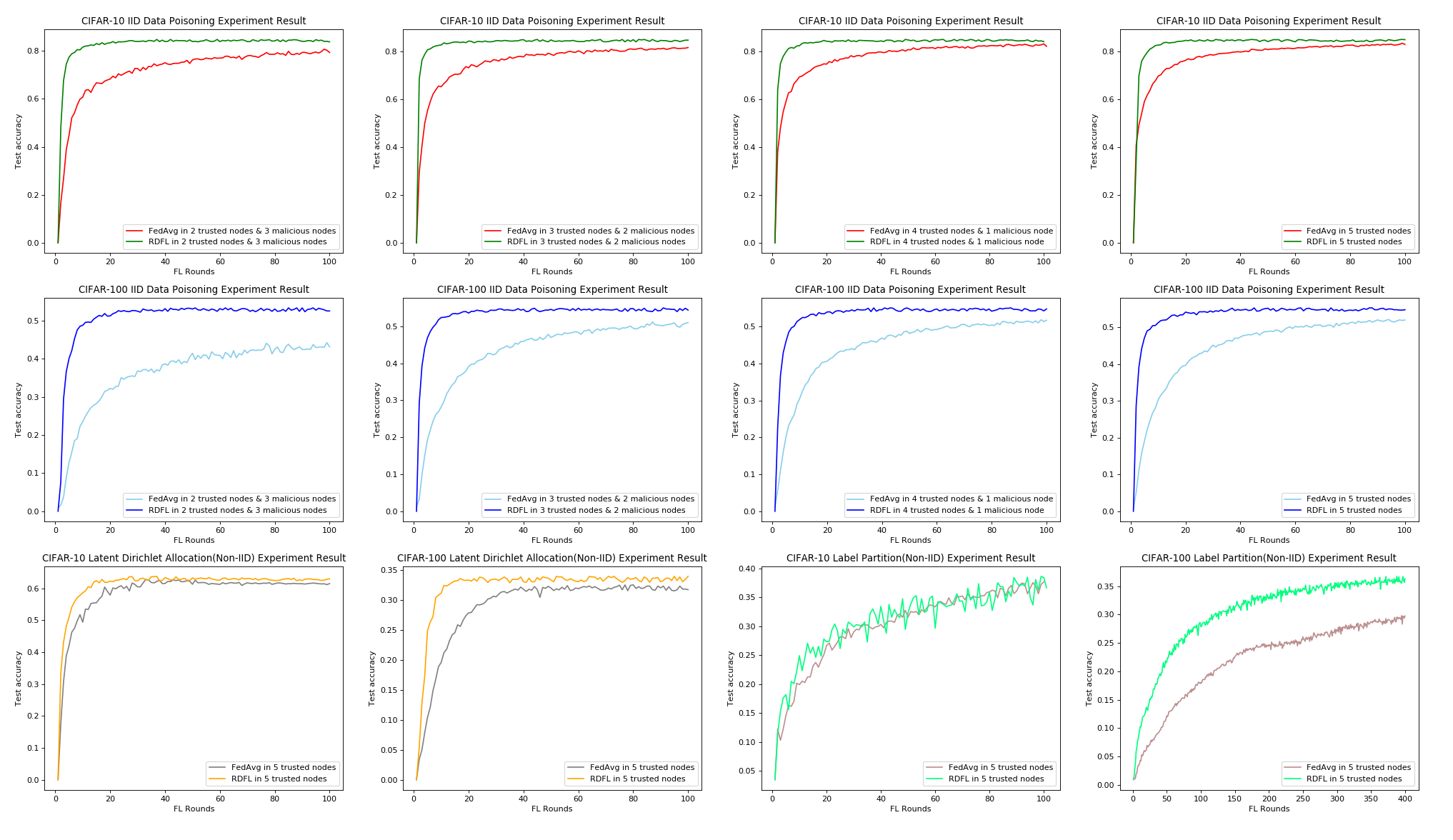}
    \caption{RDFL Experiment Result}
    \label{fig:rdfl_result}
\end{figure*}

\begin{table*}[htb]
\centering

\caption{Comparison of accuracy between RDFL and FedAvg}
\begin{tabular}{|c|c|cc|cc|}
\hline
\multirow{2}{*}{Experiments} &
  \multirow{2}{*}{\begin{tabular}[c]{@{}c@{}}Data Node Allocation\\ (Trusted : Malicious)\end{tabular}} &
  \multicolumn{2}{c|}{FedAvg} &
  \multicolumn{2}{c|}{RDFL(ours)} \\ \cline{3-6} 
 &
   &
  \multicolumn{1}{l}{CIFAR-10} &
  \multicolumn{1}{l|}{CIFAR-100} &
  \multicolumn{1}{l}{CIFAR-10} &
  \multicolumn{1}{l|}{CIFAR-100} \\ \hline
\multirow{4}{*}{Data Poisoning(IID)} & 2:3 & 79.19 & 43.27 & \textbf{84.25} & \textbf{53.86} \\
                                     & 3:2 & 80.68 & 50.43 & \textbf{84.86} & \textbf{54.62} \\
                                     & 4:1 & 82.32 & 51.48 & \textbf{84.71} & \textbf{54.53} \\
                                     & 5:0 & 82.76 & 52.38 & \textbf{85.14} & \textbf{55.28} \\ \hline
Non-IID(LDA)                         & 5:0 & 61.72 & 32.29 & \textbf{63.25}          & \textbf{33.85}          \\
Non-IID(Label Partition)                       & 5:0 & 37.18     & 29.46     & \textbf{37.51}              & \textbf{36.29}              \\ \hline
\end{tabular}
\label{tbl:rdfl_fedavg}
\end{table*}
The architecture of GFL is composed of IPFS and blockchain. For the blockchain, GFL utilizes Ethereum \cite{wood2014ethereum} as the implementation of blockchain. In GFL, the blockchain consists of all data nodes and the IPFS is composed of trusted nodes. we can know that trusted data nodes are the nodes that constitute the IPFS and blockchain, and the untrusted nodes are only the nodes that constitute the blockchain.

Smart contracts are protocols deployed on the blockchain through which the blockchain can perform more functions than digital transactions. In GFL, data nodes send the model's IPFS hash to the blockchain in the form of transaction by calling smart contract function. The blockchain first executes the consensus mechanism to verify whether the transaction has been tampered and if the verification is passed, the transaction will be packaged into block. 

As shown in Figure \ref{fig:gfl_architecture}, there are two types of smart contracts in GFL. The controller contract is responsible for controlling the flow of all FL tasks. The storage contract is responsible for storing the IPFS hash of the model and each FL task has an independent storage contract.

\begin{table*}[htb]
\centering
\caption{Comparison of time-consuming communication between data nodes}
\begin{tabular}{|c|c|c|c|c|c|c|c|c|c|}
\hline
\multirow{2}{*}{ModelType} &
  \multicolumn{3}{c|}{Conventional Decentralized FL} &
  \multicolumn{3}{c|}{GFL(IPFS only)} &
  \multicolumn{3}{c|}{GFL(IPFS \& blockchain)} \\ \cline{2-10} 
 & 5(r) & 10(r) & 50(r) & 5(r) & 10(r) & 50(r) & 5(r) & 10(r) & 50(r) \\ \hline
LeNet-5 &
  0.24s &
  0.49s &
  2.45s &
  \textbf{0.11s} &
  \textbf{0.16s} &
  \textbf{0.58s} &
  \textbf{0.41s} &
  \textbf{0.43s} &
  \textbf{0.93s} \\
ResNet-50 &
  79.63s &
  161.54s &
  796.90s &
  \textbf{15.92s} &
  \textbf{15.99s} &
  \textbf{16.41s} &
  \textbf{16.23s} &
  \textbf{16.34s} &
  \textbf{17.03s} \\ \hline
\end{tabular}
\label{tbl:time}
\end{table*}

\begin{table*}[htb]
\centering
\caption{Comparison of communication data volume between data nodes}
\begin{tabular}{|c|c|c|c|c|c|c|c|c|c|}
\hline
\multirow{2}{*}{ModelType} &
  \multicolumn{3}{c|}{Conventional Decentralized FL} &
  \multicolumn{3}{c|}{GFL(IPFS only)} &
  \multicolumn{3}{c|}{GFL(IPFS \& blockchain)} \\ \cline{2-10} 
 & 5(r) & 10(r) & 50(r) & 5(r) & 10(r) & 50(r) & 5(r) & 10(r) & 50(r) \\ \hline
LeNet-5 &
  1.19M &
  2.38M &
  11.91M &
  \textbf{0.48M} &
  \textbf{0.48M} &
  \textbf{0.48M} &
  \textbf{0.48M} &
  \textbf{0.48M} &
  \textbf{0.48M} \\
ResNet-50 &
  451.23M &
  902.45M &
  4512.26M &
  \textbf{180.51M} &
  \textbf{180.51M} &
  \textbf{180.51M} &
  \textbf{180.51M} &
  \textbf{180.51M} &
  \textbf{180.51M} \\ \hline
\end{tabular}
\label{tbl:data}
\end{table*}

\subsection{Encryption Mechanism}
Due to the visibility of blockchain, the nodes on blockchain could access transaction information in any block and the existence of malicious nodes in untrusted data nodes may cause data privacy leakage. To solve the problem of data privacy leakage, GFL utilizes RSA asymmetric encryption algorithm and AES symmetric encryption algorithm to encrypt the transaction information.

According to the RDFL algorithm, the data node $DP_h, h\subseteq[1,n]$ will send the model's IPFS hash to the nearest trusted node $DP_k, k\subseteq(h,n]$ in the clockwise direction. If $DP_h$ and $DP_k$ are communicating for the first time, $DP_k$ generates an AES key and encrypts it with the public key of $DP_h$ and sends it to $DP_h$ before communicating. $DP_h$ utilizes its private key to decrypt the AES key after receiving it. Then, the IPFS hashes transmitted between $DP_h$ and $DP_k$ will be encrypted with AES key.

Through the encryption mechanism of GFL, the IPFS hash of the model in each blockchain transaction is encrypted to prevent the data privacy leakage.

\section{Experiments and Results}

In this section, we first evaluate decentralized FL performance of the RDFL algorithm, then evaluate the communication time-consuming and communication data volume of GFL.

\subsection{RDFL Algorithm Analysis}

We conduct simulation experiments to verify the performance of the RDFL algorithm. The settings of the simulation experiment are as follows:

\textbf{Data Settings} We utilize CIFAR-10, CIFAR-100 \cite{krizhevsky2009learning} and MNIST \cite{lecun1998gradient} dataset to verify the performance of the RDFL algorithm. We simulate 5($n=5$) data nodes in this experiment. To verify the FL performance of RDFL under the data poisoning of malicious nodes. The MNIST dataset is simulated as a malicious dataset. We divide datasets into independent and identically distributed(IID) datasets and assign them to data nodes. In addition, to verify the FL performance of the RDFL algorithm on the Non-IID datasets, we utilize the Latent Dirichlet Allocation(LDA) \cite{he2020fedml} and label partition method to divide the datasets into Non-IID datasets. When using label partition method, each partition only includes two classes of CIFAR-10 or twenty classes of CIFAR-100 dataset.   

\textbf{Model Settings} we use a convolutional neural networks(CNN) with three 3x3 convolution layers(the first with 32 channels, the second with 64 channels and the three with 64 channels, each followed with 2x2 max pooling and ReLu activation) and two FC layers.

\textbf{Training Settings} At each data node, we use the SGD algorithm to train the CNN model mentioned above. The learning rate and weight decay are both 0.001. The batch size is 64. In each FL round, the data node performs 5 rounds of local model training.

\textbf{Comparison Settings}
We assume that all untrusted nodes are malicious nodes and set up malicious data nodes in 5 data nodes according to 4 different ratios to compare the performance of the RDFL algorithm and FedAvg. In addition, we also compared the performance of the RDFL algorithm and FedAvg under different Non-IID datasets without malicious nodes.

\textbf{Results}
To verify the decentralized FL performance of the RDFL algorithm under the data poisoning of malicious nodes, we conduct the RDFL algorithm and FedAvg under four node ratios. Figure \ref{fig:rdfl_result} shows the process of the training and Table \ref{tbl:rdfl_fedavg} shows that the RDFL algorithm has better performance than FedAvg at different node ratios.

In addition, to verify the performance of the RDFL algorithm under the Non-IID datasets, we use the Latent Dirichlet Allocation(LDA) and label partition method to divide dataset into 5 partitions to compare the performance of the RDFL algorithm and FedAvg. Figure \ref{fig:rdfl_result} shows the process of the training. It can be seen from Table \ref{tbl:rdfl_fedavg} that the RDFL algorithm has better performance on Non-IID datasets.

\subsection{Communication Analysis}
We conduct simulation experiments to verify the communication performance of the GFL. The settings of the simulation experiment are as follows:

\textbf{Experiment Settings}
We build an IPFS network and blockchain with 3 trusted data nodes(RAM:16GB; CPU:Intel Xeon Gold 5118 2.30GHz) and utilize the Lenet-5 model and the ResNet-50 model to verify the communication performance of GFL. 

\textbf{Results}
We firstly verify the communication performance of conventional decentralized FL framework \cite{he2019central}, then verify the communication performance of GFL when only IPFS is introduced. Finally we verify the communication performance of GFL when IPFS and blockchain both introduced. Table \ref{tbl:time} and Table \ref{tbl:data} show that IPFS greatly reduces communication time and communication data volume. In addition, due to the consensus algorithm of the blockchain, the communication time consumption increases slightly after GFL introduces the blockchain. In summary, GFL has better communication performance than conventional decentralized FL frameworks.

\section{Conclusion}
 In this paper, we propose a decentralized FL framework based on blockchain called GFL to tackle the problems in existing decentralized FL frameworks. GFL utilizes consistent hashing algorithm and RDFL algorithm to improve communication performance, decentralized FL performance and stability. Moreover, GFL introduces IPFS and blockchain to further improve communication performance and FL security. We hope that GFL can help more decentralized FL applications and we welcome any suggestions for improvement.

\bibliographystyle{named}
\bibliography{ijcai21}

\end{document}